\documentclass{article}

\usepackage[preprint]{neurips_2025}

\usepackage[utf8]{inputenc} 
\usepackage[T1]{fontenc}    
\usepackage{hyperref}       
\usepackage{url}            
\usepackage{booktabs}       
\usepackage{amsfonts}       
\usepackage{nicefrac}       
\usepackage{microtype}      
\usepackage{xcolor}         

\usepackage{amsmath}
\usepackage{amssymb}
\usepackage{mathtools}

\usepackage{graphicx,amsthm,wrapfig}
\usepackage{float}
\usepackage{subfigure}

\usepackage{tikz}
\usetikzlibrary{graphs,graphs.standard,arrows}
\usepackage[capitalize,noabbrev]{cleveref}

\usepackage{algorithm}
\usepackage{algorithmic}
\usepackage{threeparttable}
\usepackage{multicol}

\newtheorem{example-set}{Example}
\newtheorem{theorem}{Theorem}
\newtheorem{definition}{Definition}

\newtheorem{assumption}{Assumption}
\newtheorem{lemma}{Lemma}

\newtheorem{problem definition}{Problem Definition}

\usepackage{enumitem}

\usepackage{bbding} 

\usepackage{bm}

\usepackage{centernot}

\usepackage{pgf}
\newdimen\arrowsize
\pgfarrowsdeclare{arcsq}{arcsq}
{
	\arrowsize=0.2pt
	\advance\arrowsize by .5\pgflinewidth
	\pgfarrowsleftextend{-4\arrowsize-.5\pgflinewidth}
	\pgfarrowsrightextend{.5\pgflinewidth}
}
{
	\arrowsize=0.8pt
	\advance\arrowsize by .5\pgflinewidth
	\pgfsetdash{}{0pt} 
	\pgfsetroundjoin   
	\pgfsetroundcap    
	\pgfpathmoveto{\pgfpoint{0\arrowsize}{0\arrowsize}}
	\pgfpatharc{-90}{-140}{4\arrowsize}
	\pgfusepathqstroke
	\pgfpathmoveto{\pgfpointorigin}
	\pgfpatharc{90}{140}{4\arrowsize}
	\pgfusepathqstroke
}

\title{Confounded Causal Imitation Learning with Instrumental Variables}

\author{%
  Yan Zeng$^{1}$, Shenglan Nie$^{1}$, Feng Xie$^{1}$\thanks{Corresponding author} $\ $, Libo Huang$^{2}$, Peng Wu$^{1}$, Zhi Geng$^{1}$ \\
$^1$Beijing Technology and Business University \\
$^{2}$ Chinese Academy of Sciences, Institute of Computing Technology
}

\begin{document}

\maketitle

\begin{abstract}
Imitation learning from demonstrations usually suffers from confounding effects of unmeasured variables (i.e., unmeasured confounders) on the states and actions. If ignoring them, a biased estimation of the policy would be entailed. To break up this confounding gap, in this paper, we take the best of the strong power of instrumental variables (IV) and propose a Confounded Causal Imitation Learning (C2L) model. This model accommodates confounders that influence actions across multiple timesteps, rather than being restricted to immediate temporal dependencies.
We develop a two-stage imitation learning framework for valid IV identification and policy optimization. In particular, in the first stage, we construct a testing criterion based on the defined pseudo-variable, with which we achieve identifying a valid IV for the C2L models. Such a criterion entails the sufficient and necessary identifiability conditions for IV validity. In the second stage, with the identified IV, we propose two candidate policy learning approaches: one is based on a simulator, while the other is offline. Extensive experiments verified the effectiveness of identifying the valid IV as well as learning the policy.
\end{abstract}
  

\section{Introduction}
\label{intro}

\begin{figure}[ht]
    \centering
    \subfigure{
    \includegraphics[width = 0.32\textwidth]{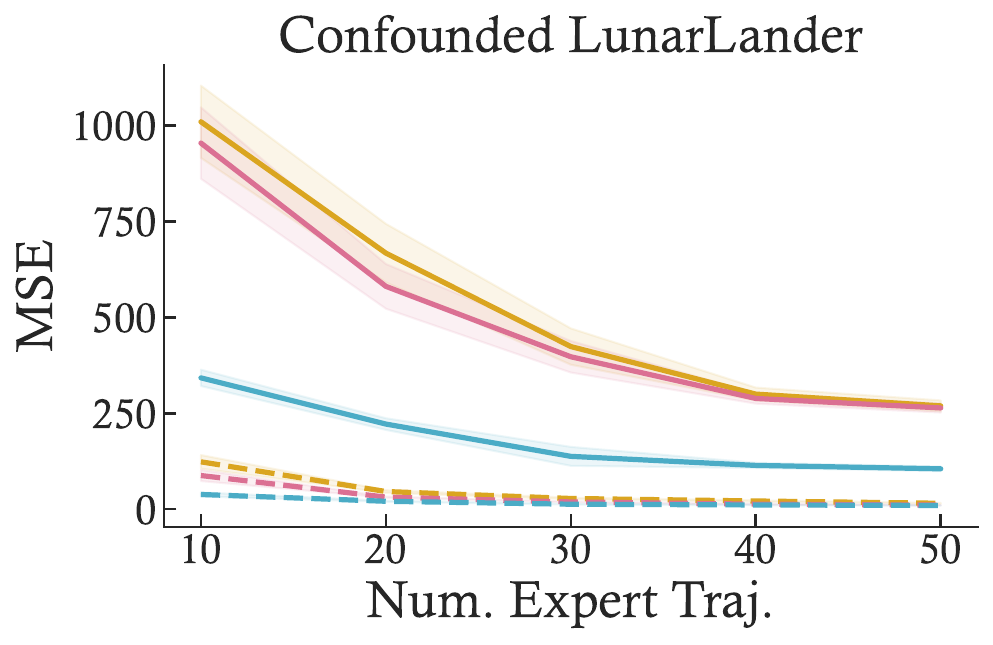}}
    \subfigure{
    \includegraphics[width = 0.32\textwidth]{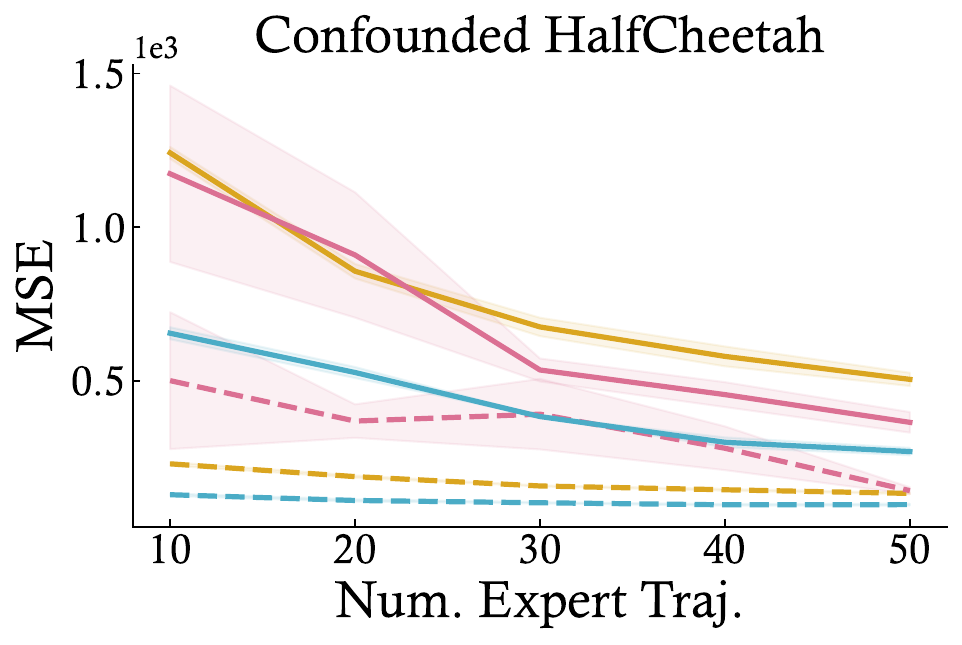}}
    \subfigure{
    \includegraphics[width = 0.32\textwidth]{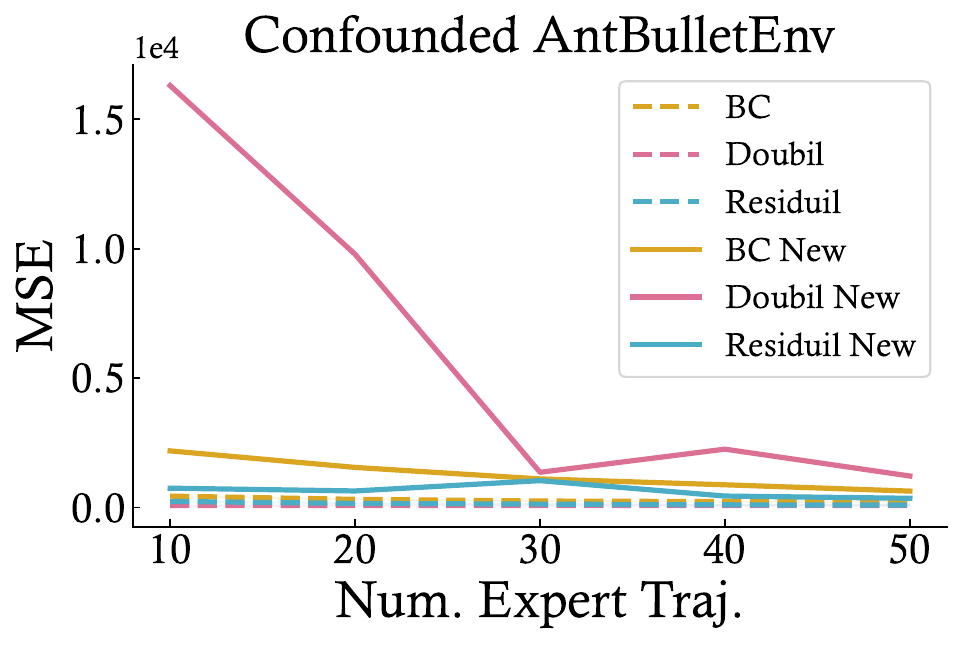}}
    
    \subfigure{
    \includegraphics[width = 0.32\textwidth]{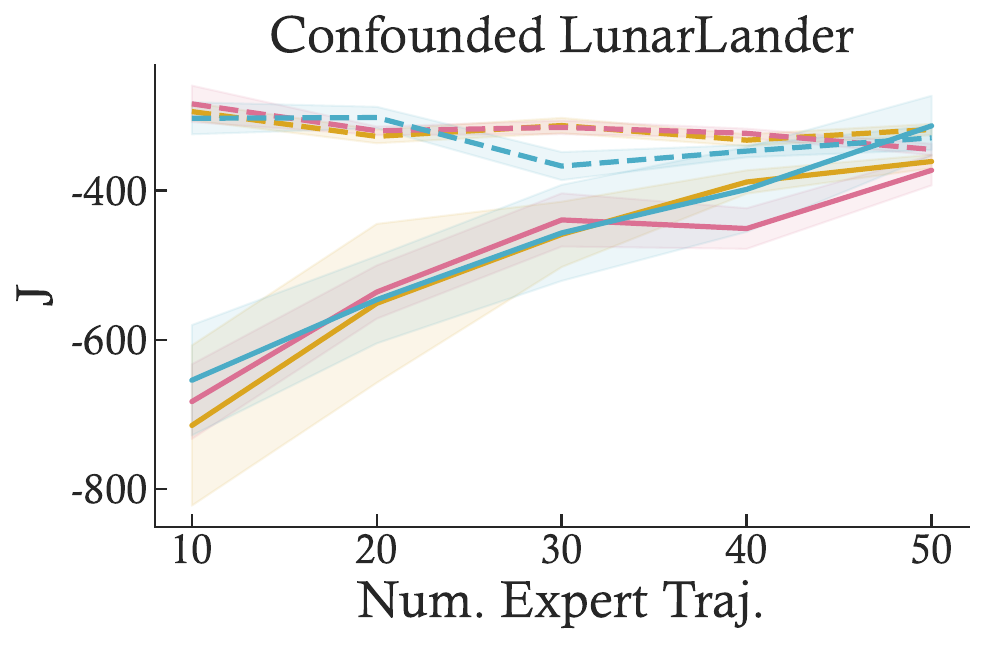}}
    \subfigure{
    \includegraphics[width = 0.32\textwidth]{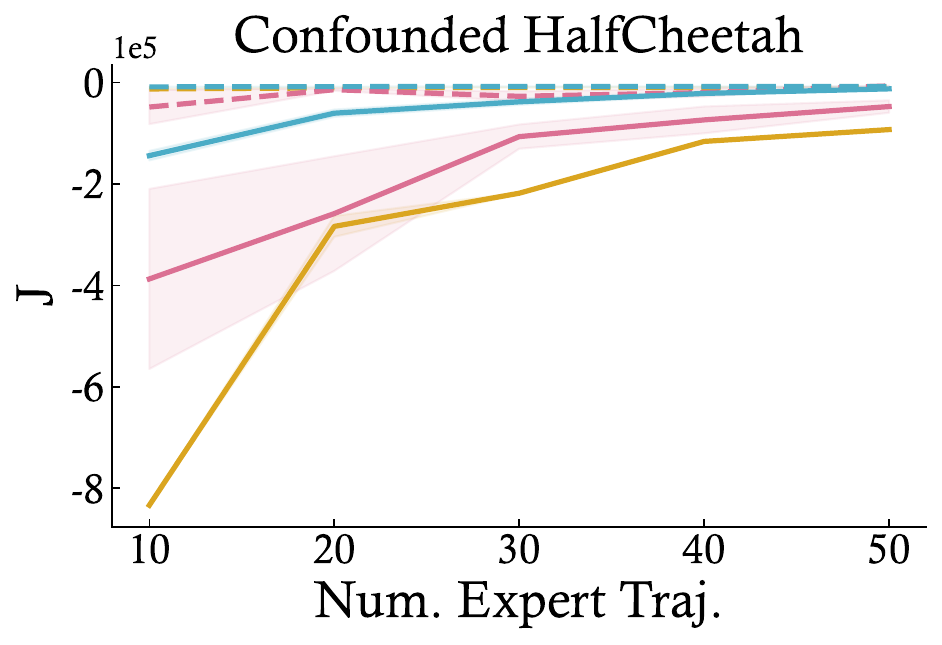}}
    \subfigure{
    \includegraphics[width = 0.32\textwidth]{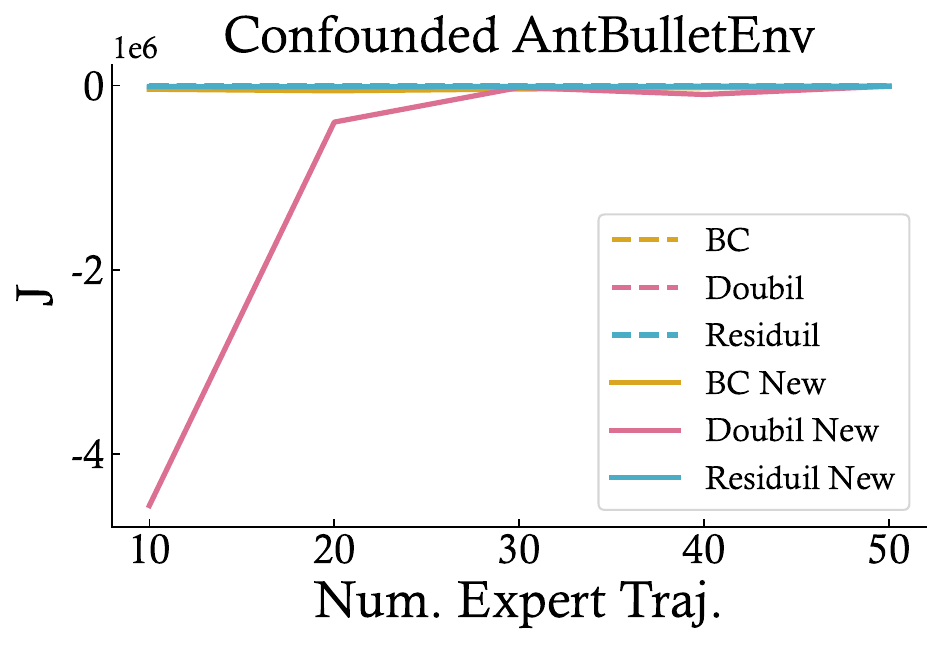}}
    \caption{Experimental results of the mean square error ($\mathrm{MSE}$) and the policy value ($\mathrm{J}$) when the generation mechanism of training demonstrations changes from single-timestep 2-TCN to multiple-timestep 3-TCN, where confounders persist across three consecutive actions, other than two.
    Methods marked with "New" indicate performance under the 3-TCN setting, while unmarked ones correspond to the original 2-TCN scenario. Behavioral Cloning (BC) is a traditional IL method.
    }
    \label{fig:mse_j in intro}
\end{figure}

Imitation learning (IL) enables autonomous agents to learn decision-making policies by mimicking expert demonstrations, offering a powerful paradigm for complex tasks where reward engineering is challenging~\citep{schaal1999imitation,ho2016generative,hussein2017imitation,zare2024survey}. 
It has been successfully applied across diverse domains, including robotics control, autonomous driving, and healthcare, etc.~\citep {van2010superhuman,zhu2018robot,le2022survey,aytar2018playing}.
While IL has achieved remarkable success, its real-world deployment faces a critical challenge, that the estimated expert policy is biased with imperfect demonstrations~\citep{ravichandar2020recent}. 

A fundamental source of such bias stems from unmeasured confounders.
These confounders are those latent variables that simultaneously influence both the expert's observed states and actions, corrupting the learned policy via confounding bias. 
They pervade real-world application scenarios. For instance, in autonomous driving, unobserved driver fatigue or environmental distractions (e.g., phone use) systematically influence both vehicle states (e.g., speed) and actions (e.g., steering)~\citep{oviedo2016understanding}; in quadcopter flying, the persistent wind could serve as a latent confounder to affect the quadcopter's position as well as its heading~\citep{kim2003autonomous}.
These ubiquitous cases demonstrate that conventional IL methods, which ignore confounding, inevitably learn policies with bias. Consequently, developing confounder-robust IL frameworks becomes not just theoretically interesting but practically essential.

Recent work has begun addressing the confounding problem in IL through the lens of causal inference.
%
Most relevantly, \citet{swamy2022causal} built on instrumental variable (IV) methods to handle confounding. They proposed Temporally Correlated Noise (TCN) models that utilize the immediate past state 
as an IV, and developed two efficient algorithms, namely DoubIL and ResiduIL, for policy learning.
While their approaches successfully eliminate confounding bias, the core limitation lies in the restrictive temporal assumption that each confounder can only affect two adjacent actions, i.e., the confounding is limited to a single timestep. 
However, in realistic scenarios where confounders persist across multiple timesteps, e.g., a driver's distraction spanning several traffic interactions, the validity of past state as an IV breaks down. 
We conducted a toy experiment to examine the robustness of existing methods when their fundamental temporal assumptions are violated. Specifically, we modified the demonstration generation process to allow confounders to influence three consecutive actions (rather than two as in standard benchmarks). As evidenced in Fig.\ref{fig:mse_j in intro}, this violation leads to significant performance degradation.
These results underscore the practical limitations of approaches that cannot handle persistent multi-timestep confounding effects.

To overcome these limitations, we propose a \textbf{C}onfounded \textbf{C}ausal Imitation \textbf{L}earning (abbreviated as C2L) model, which allows confounders to influence arbitrary-length action sequences, capturing real-world persistence effects. In this model, the past state may not serve as valid IVs due to multi-timestep confounding effects, making the identification of valid IVs a fundamental challenge. As such, we first establish theoretical foundations for IV identification before proposing a two-stage causal imitation learning framework. Specifically, the contributions of our work are three-fold:
\begin{itemize}[leftmargin=15pt,itemsep=0pt,topsep=0pt,parsep=0pt]
    \item [1.] We derive necessary and sufficient conditions for valid IV identification through the defined auxiliary residual variable, demonstrating that effective IVs can indeed be identified from purely observational data. 
    Such conditions theoretically guarantee IV validity while remaining practically computable.
    \item [2.] Based on the identification theory, we develop a two-stage causal imitation learning framework with practical implementations. The first stage employs conditional independence tests to estimate valid IVs from observational data, while the second stage offers dual policy optimization approaches, one utilizing simulator access and another operating purely offline, to learn the imitation policy.
    \item [3.] Extensive experiments across three distinct environments (LunarLander, HalfCheetah, and AntBulletEnv) validate both the accuracy of our IV identification and the superior policy learning performance compared with the baseline. The results consistently demonstrate our method's robustness under varying confounding durations and distributions. 
\end{itemize}

\section{Problem formalization}
\subsection{Preliminaries}
Before giving our model's definition, we first provide concepts of latent confounders, instrumental variables (IV), and IV estimator.

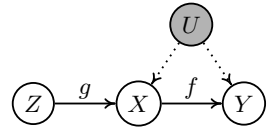
\begin{wrapfigure}{r}{0.44\textwidth} 
        \centering
	\begin{tikzpicture}[scale=0.7, line width=0.7pt, inner sep=0.5mm]
	\draw (-2, 1) node(Z) [circle, minimum size=0.5cm, draw] {{\footnotesize\,$Z$\,}};	
        \draw (0, 1) node(X) [circle, minimum size=0.5cm, draw] {{\footnotesize\,$X$\,}};
		\draw (2, 1) node(Y) [circle, minimum size=0.5cm, draw] {{\footnotesize\,$Y$\,}};
		\draw (1, 2.5) node(U) [circle, fill=gray!60, minimum size=0.5cm, draw] {{\footnotesize\,$U$\,}};

		\draw[-arcsq] (X) -- (Y) node[pos=0.5,sloped,above] {{\small\,$f$}\,};
		\draw[-arcsq] (Z) -- (X) node[pos=0.5,sloped,above] {{\small\,$g$}\,};
		\draw[-arcsq,dotted] (U) -- (X) node[pos=0.5,sloped,above] {};
		\draw[-arcsq,dotted] (U) -- (Y) node[pos=0.5,sloped,above] {}; 
		\end{tikzpicture}
	\caption{A graphical illustration of the valid IV model, where the unshaded variables are observed while the shaded one $U$ is unobserved. Here $U$ is a latent confounder and $Z$ is a valid IV. To consistently estimate the causal effects (derived by $f$) from $X$ to $Y$, we can utilize the valid IV $Z$ to eliminate the unobserved effects from $U$.} 
	\label{fig:IV_conf}
	\vspace{-25mm}
\end{wrapfigure}

A \textbf{\textit{latent confounder}} in a causal graph is the unobserved direct common cause of two observed variables. 
For example, suppose we have random variables $X, Y$, and $Z$ on the sample space $\mathcal{X}, \mathcal{Y}$, and $\mathcal{Z}$. Assume that $X, Y$, and $Z$ follow the causal structure in Fig.\ref{fig:IV_conf}, where the unshaded variables $X, Y$, and $Z$ are observed while the shaded one $U$ is unobserved. 
Here $U$ is a latent confounder.
Especially, confounders in the potential outcome framework are those variables that affect both the treatment $X$ and the outcome $Y$~\cite{glymour2019review,Pearl2000causality}. Latent confounders are ubiquitous in many real-world scenarios. Ignoring them when identifying the causal relationships between the treatment and the outcome, would induce inconsistent estimates.


To deal with these latent variables, the instrumental variable model is one of the canonical yet effective techniques~\cite{angrist1996identification}. One can use the IV to explore the causal effects of the treatment $X$ on the outcome $Y$ from observational data. 
As illustrated in Fig.\ref{fig:IV_conf}, $Z$ is a valid IV.
Formally, the \textit{\textbf{valid IV}} $Z$ must satisfy the following three assumptions~\cite{wright1928tariff,bowden1990instrumental,hernan2006instruments}:
\begin{itemize}
    \item [A1.] [\emph{Relevance}] $Z$ is associated with the treatment $X$;
    \item [A2.]  [\emph{Exclusion Restriction}] $Z$ has no direct path to the outcome $Y$;
    \item [A3.] [\emph{Exogeneity}] $Z$ is independent with the latent confounder $U$.
\end{itemize}

Given the valid IV, one can apply the classical \textit{\textbf{IV estimator}}, the two-stage
ordinary least squares (TSLS) estimator, to consistently estimate the causal effects of interest~\cite{angrist1996identification}. TSLS first regresses $X$ on $Z$ to construct the predicted $\hat{X}$ that is not correlated with $U$, and then regresses $Y$ on $\hat{X}$, returning the estimated causal effects from $\hat{f}$. Intuitively, the IV $Z$ that is independent of $U$, introduces a source of exogenous variation (randomness), which "cleans" $X$ of the confounding bias caused by $U$, enabling an unbiased estimate of the causal effect of $X$ on $Y$.

\subsection{Confounded causal imitation learning models}
We propose Confounded Causal Imitation Learning Models (abbreviated as C2L models), followed by notations.

We focus on a Markov Decision Process (MDP) defined by the tuple $< \mathcal{S}, \mathcal{A}, \mathcal{T}, r, T >$, where $\mathcal{S}$ denotes the state space, $\mathcal{A}$ is the action space, $\mathcal{T}: \mathcal{S} \times \mathcal{A} \to 
\Delta(S)$ is the transition operator that specifies the conditional distributions over states $S$, $\Delta(S)$ is the set of distributions over $S$, $r: \mathcal{S} \times \mathcal{A} \to \lbrack -1, 1\rbrack$ is the reward function, and $T$ is the time horizon of the trajectory.
Denote by $\pi: \mathcal{S} \to \Delta(A)$ the 
policy, $\Delta(A)$ is the set of distributions over actions $A$, and $\Pi \in 
\lbrace \mathcal{S} \to \Delta(A) \rbrace$ be the policy class that we optimize over.
Let $J(\pi) = \mathbb{E}_{\pi} \lbrack \sum_{t=0}^T r(s_t, a_t)\rbrack$ be the value of a policy $\pi$.

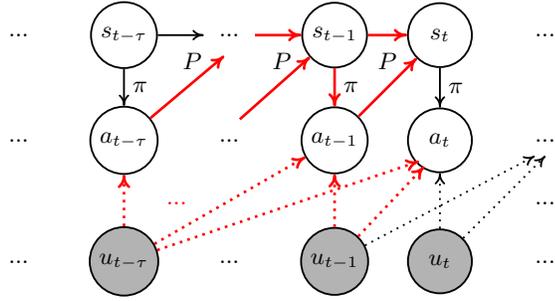
\begin{wrapfigure}{r}{0.54\textwidth} 
	\begin{center}
	\begin{tikzpicture}[scale=0.7, line width=0.7pt, inner sep=0.5mm]	
    \draw (-6, 2.5) node(s1)  {{\footnotesize\,$\; ... \;$\,}};
    \draw (4, 2.5) node(s2)  {{\footnotesize\,$\; ... \;$\,}};
    \draw (-4, 2.5) node(st-tau) [circle, minimum size=0.75cm, draw] {{\footnotesize\,$s_{t-\tau}$\,}};	
    \draw (-2, 2.5) node(st-)  {{\footnotesize\,$\; ... \;$\,}};	
    \draw (0, 2.5) node(st-1) [circle, minimum size=0.75cm, draw] {{\footnotesize\,$s_{t-1}$\,}};
	\draw (2, 2.5) node(st) [circle, minimum size=0.85cm, draw] {{\footnotesize\,$s_t$\,}};
    \draw (-2.7, 2) node(p1)  {{\footnotesize\,$P$\,}};
    \draw (-1, 2) node(p2)  {{\footnotesize\,${P}$\,}};
    \draw (1, 2) node(p2)  {{\footnotesize\,${P}$\,}};
    
        \draw (-4, 0.5) node(at-tau) [circle, minimum size=0.75cm, draw] {{\footnotesize\,$a_{t-\tau}$\,}};	
        \draw (-2, 0.5) node(at-)  {{\footnotesize\,$\; ... \;$\,}};
        \draw (0, 0.5) node(at-1) [circle, minimum size=0.85cm, draw] {{\footnotesize\,$a_{t-1}$\,}};
		\draw (2, 0.5) node(at) [circle, minimum size=0.85cm, draw] {{\footnotesize\,$a_t$\,}};
        \draw (-6, 0.5) node(-)  {{\footnotesize\,$\; ... \;$\,}};
        \draw (4, 0.5) node(--)  {{\footnotesize\,$\; ... \;$\,}};
    \draw (-4, -1.8) node(ut-tau) [circle, fill=gray!60, minimum size=0.85cm, draw] {{\footnotesize\,$u_{t-\tau}$\,}};
    \draw (-2, -1.8) node(ut-)  {{\footnotesize\,$\;...\;$\,}};
    \draw (0, -1.8) node(ut-1) [circle, fill=gray!60, minimum size=0.85cm, draw] {{\footnotesize\,$u_{t-1}$\,}};
    \draw (2, -1.8) node(ut) [circle, fill=gray!60, minimum size=0.85cm, draw] {{\footnotesize\,$u_{t}$\,}};
    \draw (-3, -0.7) node(1)  {{\footnotesize\,$\;\textcolor{red}{...}\;$\,}};
    \draw (4, -0.7) node(2)  {{\footnotesize\,$\;...\;$\,}};
    \draw (-6, -1.8) node(u1)  {{\footnotesize\,$\; ... \;$\,}};
    \draw (4, -1.8) node(u2)  {{\footnotesize\,$\; ... \;$\,}};
    
		\draw[-arcsq,color = red, line width = 1pt] (st-1) -- (st) node[pos=0.5,sloped,above] {\,};
        \draw[-arcsq] (st-tau) -- (st-) node[pos=0.5,sloped,above] {\,};
        \draw[-arcsq,color = red, line width = 1pt] (st-) -- (st-1) node[pos=0.5,sloped,above] {\,};
    \draw[-arcsq] (st-tau) -- (at-tau) node[pos=0.5,right] {{\small\,$\pi$}\,};
    \draw[-arcsq,color = red, line width = 1pt] (st-1) -- (at-1) node[pos=0.5,right] {{\small\,$\textcolor{black}{\pi}$}\,};
    \draw[-arcsq] (st) -- (at) node[pos=0.5,right] {{\small\,$\pi$}\,};
		\draw[-arcsq,color = red, line width = 1pt] (at-tau) -- (-2.1,2.1) node[pos=0.5,sloped,above] {\,};
        \draw[-arcsq,color = red, line width = 1pt] (-1.8,0.9) -- (st-1) node[pos=0.5,sloped,above] {\,};
        \draw[-arcsq,color = red, line width = 1pt] (at-1) -- (st) node[pos=0.5,sloped,above] {\,};
	
    \draw[-arcsq,dotted,color = red, line width = 1pt] (ut-tau) -- (at-tau) node[pos=0.5,sloped,above] {};
	\draw[-arcsq,dotted,color = red, line width = 1pt] (ut-tau) -- (at-1) node[pos=0.5,sloped,above] {}; 
    \draw[-arcsq,dotted,color = red, line width = 1pt] (ut-tau) -- (1.6,0.1) node[pos=0.5,sloped,above] {}; 
    \draw[-arcsq,dotted,color = red, line width = 1pt] (ut-1) -- (1.7,0) node[pos=0.5,sloped,above] {};
    \draw[-arcsq,dotted,color = red, line width = 1pt] (ut-1) -- (at-1) node[pos=0.5,sloped,above] {};
    \draw[-arcsq,dotted] (ut-1) -- (3.8, 0.2) node[pos=0.5,sloped,above] {};
    \draw[-arcsq,dotted] (ut) -- (at) node[pos=0.5,sloped,above] {};
    \draw[-arcsq,dotted] (ut) -- (4, 0.2) node[pos=0.5,sloped,above] {};
		\end{tikzpicture}
		\caption{Our proposed Confounded Causal Imitation Learning (C2L) Models, where latent confounders would affect actions for $\tau$ time steps, e.g., $u_{t-\tau}$ affects $a_{t-\tau}$, ..., $a_{t-1}$ and $a_t$. 
        Note that C2L allows different numbers of confounders. 
        Red arrows here indicate the flows of confounding effects between states $s_t$ and actions $a_t$ from multiple confounders $u_{t-\tau}, ...,$ and $u_{t-1}$, resulting in the biased estimation of policy $\pi$. 
        Thus to eliminate such confoundings, we have to find which $s_{t-\tau}$ is the valid IV.
        }  
        \vspace{-5mm}
	\label{fig:C2L_model}
	\end{center}
\end{wrapfigure}

Our C2L model is depicted in Fig.$\ref{fig:C2L_model}$, which uncovers the underlying mechanisms of trajectory generation. 
$u_{t-\tau}$ is the latent confounder at timestep $t-\tau$, which influences multiple $\tau$ actions from  timestep $t-\tau$ to $t$. $\tau$ could be various among different C2L models. 
Confounders $u_{t-\tau}, ...,$ and $u_{t-1}$ influence the current action $a_t$, while they also influence the current state $s_t$ through the past actions $a_{t-\tau}$, ..., $a_{t-1}$ and the dynamics $\mathcal{T}$. 
Due to the confounding effects between $s_t$ and $a_t$ when learning the policy, it is desirable to identify the valid IV to avoid biased estimation of $\pi$.
Specifically, the generation mechanisms of current states and actions satisfy the following structural causal models.
\begin{equation}\label{equ:C2Lmodel}
\begin{aligned}
    s_t & = P(s_{t-1}, a_{t-1})+e_{s_t}, \\
    & = P(s_{t-1}, \pi(s_{t-1}) + h(u_{t-1}, ...,u_{t-\tau},u_{t-\tau-1})+e_a) +e_{s_t},\\
    a_t & = \pi(s_t) + h(u_t, u_{t-1},...,u_{t-\tau})+e_{a_t},\\
\end{aligned}
\end{equation}
where $h$ is a deterministic function describing the effects of multiple latent confounders. $e_s$ and $e_a$ are statistically independent noise terms. We see that $\tau$ latent confounders, i.e., $u_{t-1},...,u_{t-\tau}$, produce confounding effects towards $s_t$ and $a_t$. In this model, $s_{t-1}, ...,$ or $s_{t-\tau+1}$ serves as invalid IVs, since they all violate Assumption A3 \textit{Exogeneity}. $s_{t-\tau}$ can be the valid IV, satisfying all three Assumptions A1-A3 to combat the confounding effects from $u_{t-1},...,u_{t-\tau}$. Hence, with the unknown parameter $\tau$, how to pick out the valid IV $s_{t-\tau}$ becomes the major challenge.

Notably, the TCN (Temporally Correlated Noise) model~\cite{swamy2022causal} differs from our model in several key aspects. First, we allow the confounder to influence multiple actions, rather than being limited to only two consecutive actions. This implies that, in policy learning, the sources of confounding effects are more diverse, making it more challenging to eliminate such effects. Second, our model operates under the assumption that the causal model is unknown in advance, meaning that we need to automatically identify which historical states serve as valid IVs, rather than the default $s_{t-1}$. 
The TCN model can be regarded as a special case of our model. Specifically, when $\tau=2$ and all noise terms are set to zero, our model reduces to TCN.

\textbf{Our goal} is to devise a theoretical and practical causal imitation learning framework to deconfound the confounding effects from latent confounders on the policy, taking the best of the IV variable theory.
Since different numbers of confounders could affect the policy in various situations, the valid IV for consistent policy estimation could also be different. Thus, how to find a valid IV from purely observational data and entail the debiased policy estimation becomes the major challenge.

\section{Identifiability of the valid IV}

We derive an Auxiliary-Based testing Criterion (abbreviated as AB Criterion) for the IV validity, which defines an auxiliary residual variable and constrains the independent relationship between the auxiliary and a candidate IV. Such independence characteristics on the IV validity have not been achieved in imitation learning to tackle the latent confounding problem.

\begin{definition}[Auxiliary-Based testing Criterion, namely AB Criterion] \label{def:auxiliary}
    Suppose the current state $s_t$, action $a_t$, and a candidate past state $s_k$ at time step $k$ are variables in a causal model. Define the auxiliary residual variable $\mathcal{R}$ as
    \begin{equation}
        \mathcal{R}_{s_t, a_t || s_k} := a_t - l(s_t),
    \end{equation}
    where $l(\cdot)$ satisfies $\mathbb{E}[\mathcal{R}_{s_t, a_t || s_k} \mid s_k ]   =0$ and $l(\cdot)\neq 0$. We say that $\lbrace s_t, a_t || s_k \rbrace$ satisfies the AB Criterion if and only if $\mathcal{R}_{s_t, a_t || s_k} $ is independent of $s_k$.
\end{definition}

In the following, we give the necessary and sufficient conditions to test IV validity with the AB Criterion for our C2L model. Please note that all proofs are provided in the Appendix. 

\begin{theorem}[Necessary Condition for IV Validity] \label{theo:nece}
    Let $s_t$, $a_t$, and $s_k$ be the current state, current action, and the state at time step $k$, respectively, in Figure~\ref{fig:C2L_model}. 
    Suppose that the sample size $n\to \infty$ holds, and further the probability densities
    $s_k$ and $u_*$ are twice differentiable and positive on $(-\infty, \infty)$. 
    If $s_k$ is a valid IV relative to $s_t\to a_t$, then $\lbrace s_t, a_t || s_k \rbrace$ always satisfies the AB Criterion.
\end{theorem}
Theorem~\ref{theo:nece} indicates that if  $\lbrace s_t, a_t || s_k \rbrace$ violates the AB Criterion, then $s_k$ is an invalid IV relative to the policy. Given the necessary conditions, one may wonder what the sufficient conditions are for IV validity. To end it, we first analyze linear models, where $P, h$ and $\pi$ in Eq.\eqref{equ:C2Lmodel} represent linear functions and Eq.\eqref{equ:C2Lmodel} can be expressed as Eq.\eqref{equ:linear}. 
\begin{equation}\label{equ:linear}
\begin{aligned}
    s_t & = \rho_{s_{t-1}, s_t} s_{t-1} + \rho_{a_{t-1}, s_t} a_{t-1} +e_{s_t},\\
    a_t & =  \rho_{s_t,a_t} s_t + \rho_{u_t,a_t} u_t + ...+ \rho_{u_{t-\tau},a_t} u_{t-\tau} +e_{a_t},\\
\end{aligned}
\end{equation}
where $\rho_{\star, \bullet}$ is the direct effect of $\star \to \bullet$.
For linear models, we introduce the following partial non-Gaussianity assumption for identifying valid IVs.

\begin{assumption} [Partial Non-Gaussianity] \label{ass:1}
At least one of the following conditions holds: (i) at least one latent confounder $u$ follows a non-Gaussian distribution and affects $s_t$; ii) the noise term $e_s$ of states $s_k$ follows a non-Gaussian distribution.
\end{assumption}

The essence of Assumption~\ref{ass:1} lies in its constraint on the noise terms, requiring them to follow non-Gaussian distributions—a condition frequently encountered in real-world applications due to the Cramér Decomposition Theorem~\citep{cramer1962random}. This assumption enables the exploitation of higher-order statistical information to address the IV validity problem from the second-order information. Specifically, Lemma~\ref{lemma:linear_gau} underscores the insufficiency of relying solely on second-order statistics, as they fail to distinguish invalid IVs violating exogeneity when the noise terms are Gaussian. In contrast, Lemma~\ref{lemma:linear_nongau} demonstrates that such higher-order information can be leveraged to identify valid IVs by detecting dependencies obscured in Gaussian settings. Together, these results highlight the critical role of non-Gaussianity in IV identification, bridging a theoretical gap left by traditional linear Gaussian models in causal imitation learning. 

\begin{lemma} [Non-Testability in Linear Gaussian C2L] \label{lemma:linear_gau}
    Let $s_t$, $a_t$, and $s_k$ be the current state, current action, and the state at time step $k$, respectively, in C2L model of Figure~\ref{fig:C2L_model}.  
    Suppose that the sample size $n\to \infty$ holds.
    If all action and state variables have Gaussian-distributed noise terms, regardless of $s_k$'s IV validity, $\lbrace s_t, a_t || s_k \rbrace$ satisfies the AB Criterion.
\end{lemma}

\begin{lemma} [Testability in Linear Non-Gaussian C2L] \label{lemma:linear_nongau}
    Let $s_t$, $a_t$, and $s_k$ be the current state, current action, and the state at time step $k$, respectively, in C2L model of Figure~\ref{fig:C2L_model}.  
    Suppose that the sample size $n\to \infty$ and further Assumption~\ref{ass:1} hold. 
    If $s_k$ is an invalid IV relative to $s_t \to a_t$, $\lbrace s_t, a_t || s_k \rbrace$ violates the AB Criterion.
\end{lemma}

Theorem~\ref{theo:nece}, Lemmas~\ref{lemma:linear_gau} and~\ref{lemma:linear_nongau} induce the following theorem, which describes the necessary and sufficient conditions for IV validity in linear C2L models. 

\begin{theorem} [Necessary and Sufficient Conditions for IV Validity in Linear C2L] \label{theo:linear}
    Let $s_t$, $a_t$, and $s_k$ be the current state, current action, and the state at time step $k$, respectively, in C2L model of Figure~\ref{fig:C2L_model}.  
    Suppose that the sample size $n\to \infty$ and the probability densities
    $s_k$ and $u_*$ are twice differentiable and positive on $(-\infty, \infty)$.
    Further, suppose that Assumption~\ref{ass:1} holds. $\lbrace s_t, a_t || s_k \rbrace$ satisfies the AB Criterion if and only if the candidate IV $s_k$ is valid.
\end{theorem}

While linear C2L models provide a tractable framework for IV validity, the inherent simplicity may limit their applicability in real-world scenarios where relationships are often nonlinear. To address this critical gap, we now extend our discussion to nonlinear settings. This generalization necessitates the introduction of Assumption~\ref{ass:2}, which serves as the foundation for our nonlinear identification theorem.

\begin{assumption}[Non-Degenerate Cross-Derivative] \label{ass:2}
    Assume the second-order partial derivative of the logarithm joint density of $\mathcal{R}_{s_t, a_t||s_k}$ and $s_k$ for $\pi(\cdot)\neq l(\cdot)$    
    is non-zero, i.e.,
    $\frac{\partial^2\log p(\mathcal{R}_{s_t, a_t||s_k}, s_k)}{\partial\mathcal{R}_{s_t, a_t||s_k} \partial s_k} \neq0$,
    where $\mathcal{R}_{s_t, a_t||s_k} = \pi(s_t)-l(s_t)+h(u_t,u_{t-1},...,u_{t-\tau})+e_{a_t}$ is the estimated auxiliary residual variable, and $p(\mathcal{R}_{s_t, a_t||s_k}, s_k)$ is the joint density of $\mathcal{R}_{s_t, a_t||s_k}$ and $s_k$.
\end{assumption}

\begin{theorem} [Necessary and Sufficient Conditions for IV Validity in Nonlinear C2L]
    Let $s_t$, $a_t$, and $s_k$ be the current state, current action, and the state at time step $k$, respectively, in our C2L model of Figure~\ref{fig:C2L_model}.
    Suppose that the sample size $n\to \infty$ holds and the probability densities
    $e_s$ and $u_*$ are twice differentiable and positive on $(-\infty, \infty)$. Further, suppose that Assumption~\ref{ass:2} holds. Then $\lbrace s_t, a_t || s_k \rbrace$ satisfies the AB Criterion if and only if the candidate $s_k$ is valid.
\end{theorem}

\section{Confounded causal imitation learning framework} \label{sec:frame}
Based on the above theoretical results, we propose a two-stage framework, namely C2L (\textbf{C}onfounded \textbf{C}ausal imitation \textbf{L}earning), to eliminate the confounding effects and to learn the policy. In Stage I, we aim to seek out a valid IV from past states from observational trajectories, while in Stage II, we learn the optimal policy with the identified valid IV. The whole procedure is summarized in Algorithm~\ref{alg:framework}.

\begin{algorithm}[H]
\caption{C2L Framework}
\label{alg:framework}
\begin{algorithmic}[1]
    \REQUIRE
    Trajectory dataset $D_E=\{s_t, a_t \}_{t=0}^T$ from the expert, policy class $\Pi$, the maximum number of candidate IVs $w$, the significance level $\alpha$, and the simulator $\widehat{\mathcal{T}}$ (optional).
    \STATE Valid IV $s_k$ = FindValidIV($D_E, w, \alpha$);   {$\hfill \triangleright$ {\emph{Stage I}} }
    \STATE Policy $\pi$ = LearnPolicy($D_E,  s_k, \Pi, \widehat{\mathcal{T}}$) or LearnPolicy*($D_E, s_k, \Pi $);{$\hfill \triangleright$ \emph{Stage II}}
    \ENSURE
    The Valid IV $s_k$ and policy $\pi$.
\end{algorithmic}
\end{algorithm}

\subsection{Stage I: Finding the valid IV}
In Stage I, we propose a data-driven approach to identify a valid IV from the expert state-action transitions $D_E$. To enhance effectiveness in finite-sample settings, particularly when the number of candidate IVs is large, we introduce two key parameters: \(w\) as the maximum candidate IVs to be detected, and the significance level \(\alpha\)  for the independence tests.  
In practice, we check for the valid IV with the HSIC test, a Hilbert-Schmidt Independence Criterion-based test~\citep{zhang2018large}.
The core procedure is outlined in Algorithm~\ref{alg:validIV}. 

\begin{algorithm}[H]
\caption{FindValidIV}
\label{alg:validIV}
\begin{algorithmic}[1]
    \REQUIRE
    Trajectory dataset $D_E$, the maximum number of candidate IVs $w$, and significance level $\alpha$.
    \STATE Initialize the candidate IV set \(\mathbf{W} \leftarrow \{s_{t-1}, \dots, s_{t-w}\}\);   
    \STATE Estimate the auxiliary residual variable \(\mathcal{R}_{s_t, a_t || s_k}\);
    \STATE Iterate through candidate IVs \( s_k \in \mathbf{W} \), testing each for independence with residual \( \mathcal{R}_{s_t, a_t || s_k} \); 
    \STATE If \( \mathcal{R}_{s_t, a_t || s_k} \perp\!\!\!\perp s_k \), return \( s_k \) as valid and terminate, otherwise remove \( s_k \) from $\mathbf{W}$ and repeat Lines 3-4.
    
    \ENSURE
    Valid IV $s_k$.
\end{algorithmic}
\end{algorithm}

\subsection{Stage II: learning the optimal policy}
Building upon the identified valid IV, one can employ different imitation learning approaches to learn the policy with no confounding bias. 
Drawing inspiration from the DoubIL and ResiduIL paradigms~\citep{swamy2022causal}, we propose two policy learning algorithms. 

The first approach leverages simulator access to decouple confounding effects through a two-phase optimization process, illustrated in Algorithm~\ref{alg:policy_simu}.
Specifically, after IV validation, it first trains an initial policy \(\pi_1\) via maximum likelihood estimation on the observed state-action pairs $(s,a)\sim D_E$.
To generate confounder-free synthetic states, the method recursively propagates the validated IV \(s_k\) through the multi-step transition distribution 
\begin{equation}
P(s_t|s_k) = \sum_{s_{k+1:t-1}} \sum_{a_{k:t-1}} \left[ \prod_{\nu=k}^{t-1} \pi_1(a_\nu |s_\nu) \widehat{\mathcal{T}}(s_{\nu+1}|s_\nu, a_\nu) \right],
\end{equation}
where marginalization over intermediate states \(s_{k+1}...s_{t-1}\) and actions \(a_{k},...,a_{t-1}\) accounts for all possible trajectories originating from \(s_k\).  
In practice, this is implemented by sampling rollouts for any $(s_k,a_t)\in D_E$ from the simulator
\(
\widetilde{s}_{k+1:t} \sim \prod_{\tau=k}^{t-1} \widehat{\mathcal{T}}(s_\tau, \pi_1(s_\tau)).
\)
Crucially, these states $\widetilde{s}_t$ and actions $a_t$ are confounder-free because 
$\widetilde{s}_t \perp\!\!\!\perp u_{t}...u_{t-\tau} \mid s_k$, 
while the observed actions maintain their confounding structure, 
$a_t = \pi(s_t) + h(u_t, u_{t-1}, \dots, u_{t-\tau}) + e_{at},$
where $u_{t}...u_{t-\tau}$ are confounders.
Finally, it trains the debiased policy \(\pi\) by minimizing the Mean Squared Error (MSE) between simulated states $\widetilde{s}_t$ and expert actions $a_t$, i.e., $\pi = \arg\min_{\pi \in \Pi} \mathbb{E}_{(\widetilde{s}_t,a_t) \sim {D}_{IV}} \left[ \| a_t - \pi(\widetilde{s}_t) \|_2^2 \right]$.
Notice that such an IV-guided resampling approach, grounded in the factored transition dynamics, generalizes DoubIL~\citep{swamy2022causal} by explicitly addressing multi-step confounding propagation.

\begin{algorithm}[H]
\caption{LearnPolicy}
\label{alg:policy_simu}
\begin{algorithmic}[1]
    \REQUIRE
    Trajectory dataset $D_E$, the identified valid IV $s_k$, policy class $\Pi$, and the simulator $\widehat{\mathcal{T}}$.
    \STATE Learn a biased policy, $\pi_{1} = \arg\min_{\pi \in \Pi} \mathbb{E}_{(s,a) \sim \mathcal{D}_E} [-\log \pi(a|s)]$;   
    \STATE Generate recursively synthetic transitions, 
    $\widetilde{s}_{k+1:t} \sim \prod_{k}^{t-1} \widehat{\mathcal{T}}(s_k, \pi_{1}(s_k))$ for any $(s_k,a_t)\in D_E$,
    where \(\widetilde{s}_{k+1}\) is drawn from \(\widehat{\mathcal{T}}(s_{k}, \pi_1(s_{k}))\) and subsequent states follow the composed transition;   
    \STATE Construct the deconfounded dataset ${D}_{{IV}} = \{(\widetilde{s}_t, a_t)\}$;
    \STATE Learn the optimal policy $\pi = \arg\min_{\pi \in \Pi} \mathbb{E}_{(\widetilde{s}_t,a_t) \sim \mathcal{D}_{\text{IV}}} \left[ \|a_t - \pi(\widetilde{s}_t)\|^2 \right]$.
    \ENSURE
    Policy $\pi$.
\end{algorithmic}
\end{algorithm}

For environments where simulator access is unavailable, we can also employ a game-theoretic approach to learn the policy, an offline variant extending the adversarial learning principles of ResiduIL~\citep{swamy2022causal}. 
We reformulate the minimax objective to exploit the identified IV $s_k$,
\begin{equation} \label{eq:loss_policy}
    \min_{\pi \in \Pi} \max_{f\in \mathcal{F}} L(\pi,f) = \min_{\pi \in \Pi} \max_{f\in \mathcal{F}} \mathbb{E}_{(s_k,s_t,a_t)}[2(a_t - \pi(s_t))f(s_k) - f(s_k)^2],
\end{equation}
where $f(\cdot)$ is the discriminator function that maps IV $s_k$ to Lagrange multipliers.
Intuitively, this objective formalizes a strategic interplay between two adversarial players with distinct roles. The policy player $\pi$ minimizes prediction errors under the worst-case scrutiny from the discriminator, effectively learning a robust model.
The discriminator player $f$ serves as an adaptive critic: the term \(2(a_t-\pi(s_t))f(s_k)\) rewards \(f\) for detecting systematic prediction errors conditioned on the instrument \(s_k\), while \(-f(s_k)^2\) acts as a regularizer to penalize excessive discriminator complexity, avoiding degenerate solutions. 
For the detailed derivation of Eq.\eqref{eq:loss_policy}, please refer to~\citep{dikkala2020minimax,swamy2022causal}.
Regarding the optimization, we employ Optimistic Mirror Descent~\citep{syrgkanis2015fast} to efficiently compute approximate Nash equilibria.



\begin{algorithm}[H]
\caption{LearnPolicy*}
\label{alg:policy_off}
\begin{algorithmic}[1]
    \REQUIRE
    Trajectory dataset $D_E$, the identified valid IV $s_k$,  the policy class $\Pi$, functional class $\mathcal{F}$, and learning rate $\eta$.
    \STATE Initialize the policy \(\pi \in \Pi\), discriminator \(f \in \mathcal{F}\), momentum terms \(\widetilde{g}_\pi \gets 0\) and \(\widetilde{g}_f \gets 0\) ;   
    \WHILE{\(\pi\) not converged}
    \STATE Compute loss $L(\pi, f) = \mathbb{E}_{(s_k, s_t, a_t) \sim \mathcal{D}_E} \left[ 2(a_t - \pi(s_t))f(s_k) - f(s_k)^2 \right]$;  
    \STATE Compute gradients $g_\pi = \nabla_\pi L(\pi, f),  g_f = \nabla_f L(\pi, f)$;
    \STATE Update optimistic mirror descents $\pi \leftarrow \pi - \eta (2g_\pi - \widetilde{g}_\pi),  f \leftarrow f + \eta (2g_f - \widetilde{g}_f)$;
    \STATE Preserve momentum terms $\widetilde{g}_\pi \gets g_\pi,  \widetilde{g}_f \gets g_f$.
    \ENDWHILE
    \ENSURE
    Policy $\pi$.
\end{algorithmic}
\end{algorithm}

\section{Experiments} \label{sec:ex}
We evaluate our approach on three benchmark environments: LunarLander-v2 from the OpenAI Gym~\citep{brockman2016openai}, HalfCheetah, and AntBulletEnv-v0~\citep{coumans2016pybullet}. Expert demonstrations are generated by corrupting PPO~\citep{schulman2017proximal}  for LunarLander (SAC~\citep{haarnoja2018soft} for HalfCheetah, and AntBulletEnv environments) policies with additive noise that combines fresh Gaussian samples and cached values from prior timesteps, creating temporal dependence that violates standard imitation learning assumptions. 
For more details on such three environments, see Appendix B.1 to B.3.
Unless otherwise specified, the default configurations are applied, i.e., the number of steps a latent confounder sticks around for is 3, and the confounders follow a Gaussian distribution. 

We conducted experiments with the following task.
\begin{itemize}[leftmargin=15pt,itemsep=0pt,topsep=0pt,parsep=0pt]
\item 
\textbf{T1: Performances on IV validity.} We evaluated the impacts of (i) varying numbers of expert trajectories, i.e., $T=\{10,20,30,40,50\}$; (ii) varying confounding durations, i.e., $\tau = \{3,4,5,6,7\}$; and (iii) different confounder distributions, including Gaussian, uniform, gamma, exponential, lognormal, beta and laplace distributions. 
\item 
\textbf{T2: Performances on policy learning.} 
We conducted experiments in both confounded and unconfounded environments to show the efficacy in policy optimization with the valid IV.
\end{itemize}



\subsection{Performances on IV validity}

\begin{figure}[t]
    \centering
    \subfigure{
    \includegraphics[width = 0.32\textwidth]{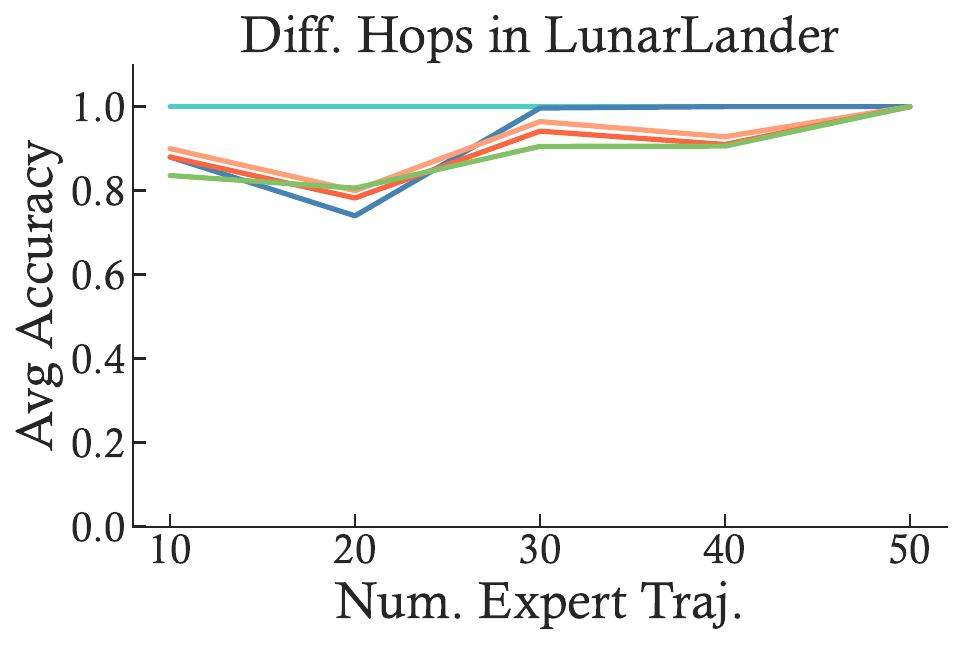}}
    \subfigure{
    \includegraphics[width = 0.32\textwidth]{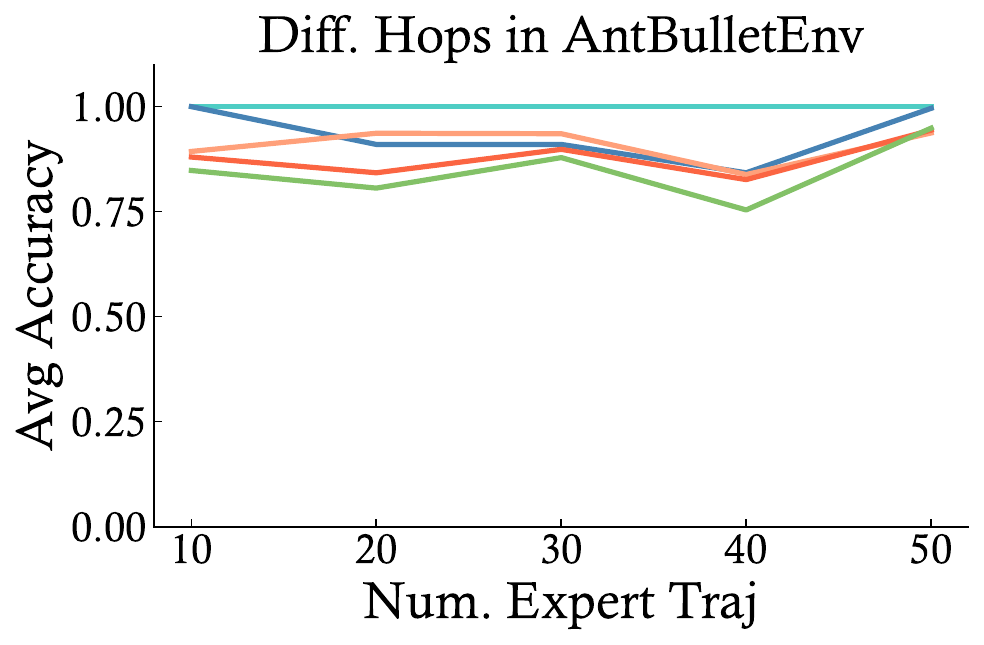}}
    \subfigure{
    \includegraphics[width = 0.32\textwidth]{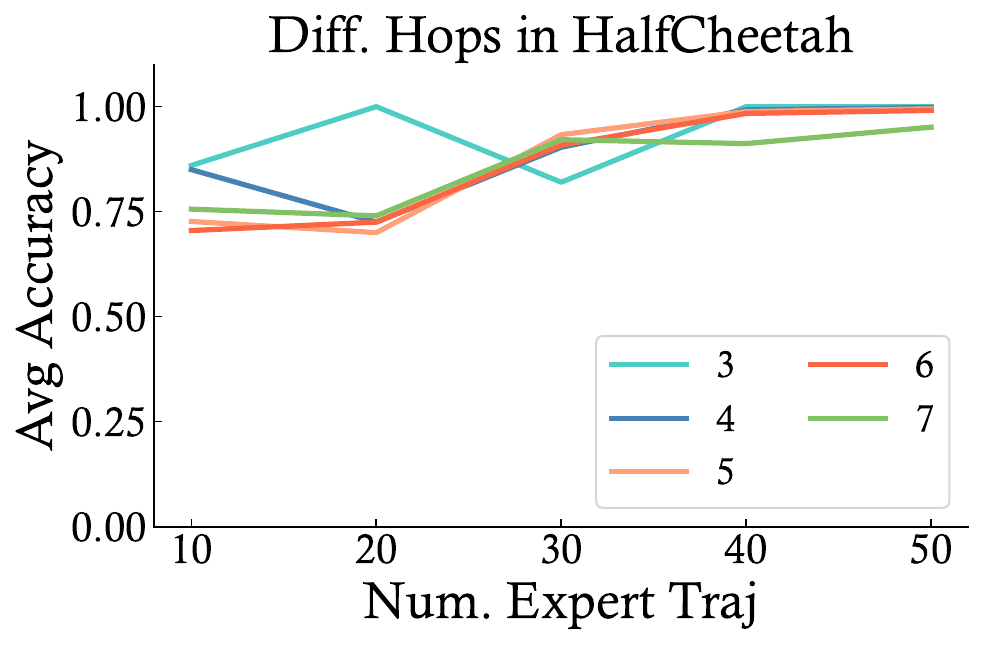}}

    \subfigure{
    \includegraphics[width = 0.32\textwidth]    {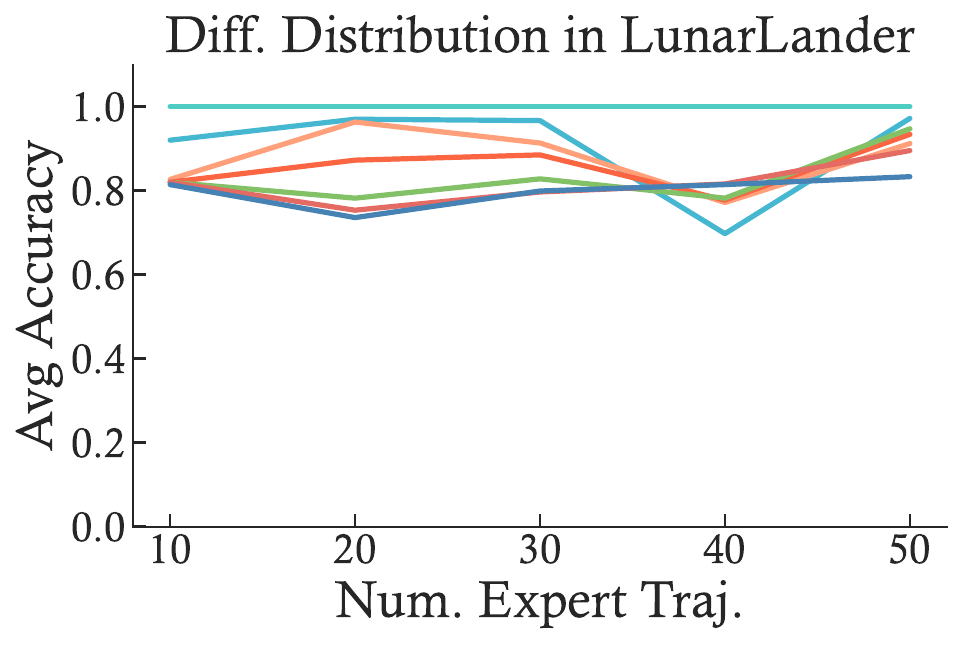}}
    \subfigure{
    \includegraphics[width = 0.32\textwidth]    {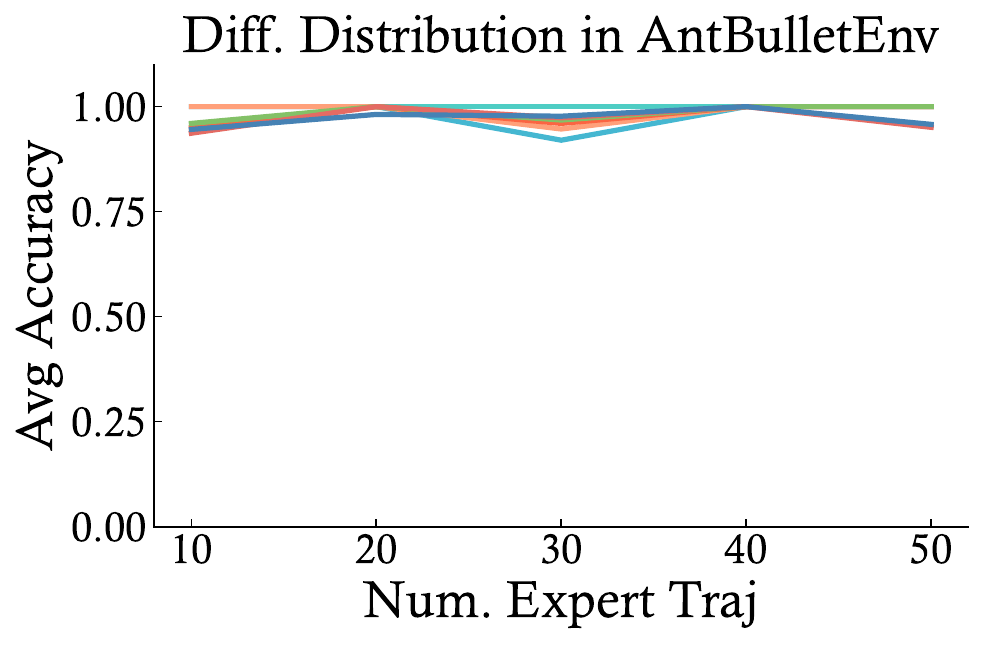}}
    \subfigure{
    \includegraphics[width = 0.32\textwidth]    {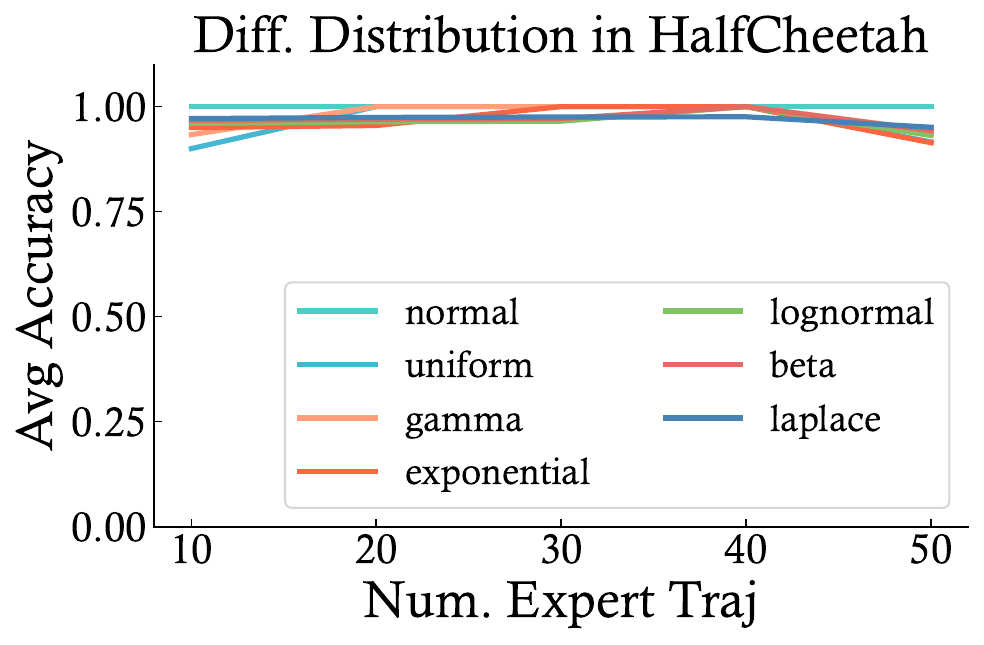}}
    \caption{Experimental results of the IV identification average accuracy in three environments, LunarLander, HalfCheetah, and AntBulletEnv. 
    Our experimental results encompass three critical aspects: (i) robustness to varying numbers of expert trajectories, i.e., $\{10,20,30,40,50\}$; (ii) robustness to varying confounding durations, i.e., $\{3,4,5,6,7\}$; and (iii) generalizability across different confounder distributions, including Gaussian, uniform, gamma, etc.
    }
    \label{fig:accuracy of valid iv identification}
    \vspace{-4mm}
\end{figure}
Here, we adopted the identification average accuracy as the evaluation metric, defined as the proportion of correctly identified IVs.
Fig.\ref{fig:accuracy of valid iv identification} demonstrates the experimental results of the IV identification average accuracy in three environments. Overall, our proposed method achieves satisfactorily convergent performances in all scenarios, implying that our AB Criterion can correctly identify valid IV in almost all cases.
Specifically, with increasing numbers of expert trajectories, it ultimately gives stable convergence across all environments. 
Regardless of the varying confounding durations of confounders, our method maintains remarkably similar convergence levels with enough trajectories, indicating the correctness of the IV identification theory. 
Notably, our method shows exceptional distributional robustness in AntBulletEnv and HalfCheetah environments, while LunarLander displays slightly greater sensitivity to non-Gaussian noise distributions.
Nevertheless, all configurations of distributions ultimately achieve satisfactory identification accuracy exceeding 80\%, with the continuous control environments particularly excelling at 90\% accuracy. 

\subsection{Performances on policy learning}

\begin{figure}
    \centering
    \subfigure{
    \includegraphics[width = 0.32\textwidth]{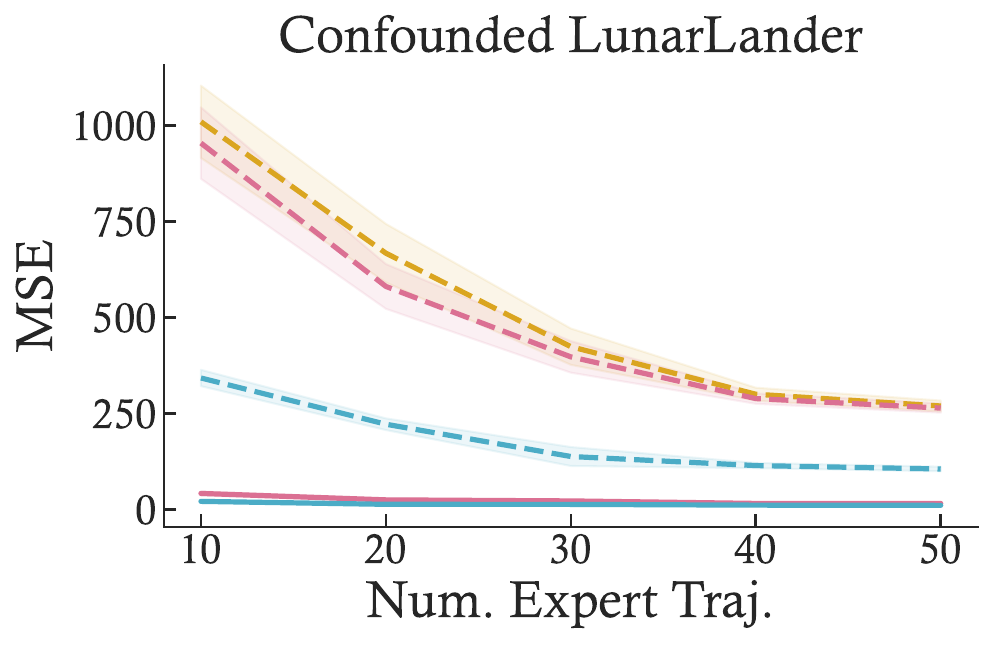}}
    \subfigure{
    \includegraphics[width = 0.32\textwidth]{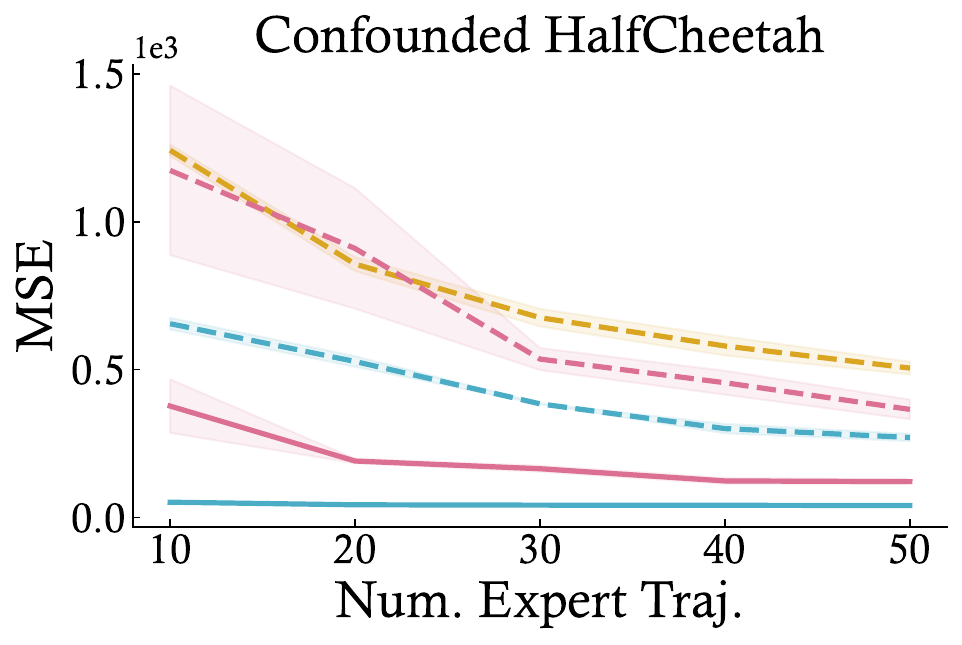}}
    \subfigure{
    \includegraphics[width = 0.32\textwidth]{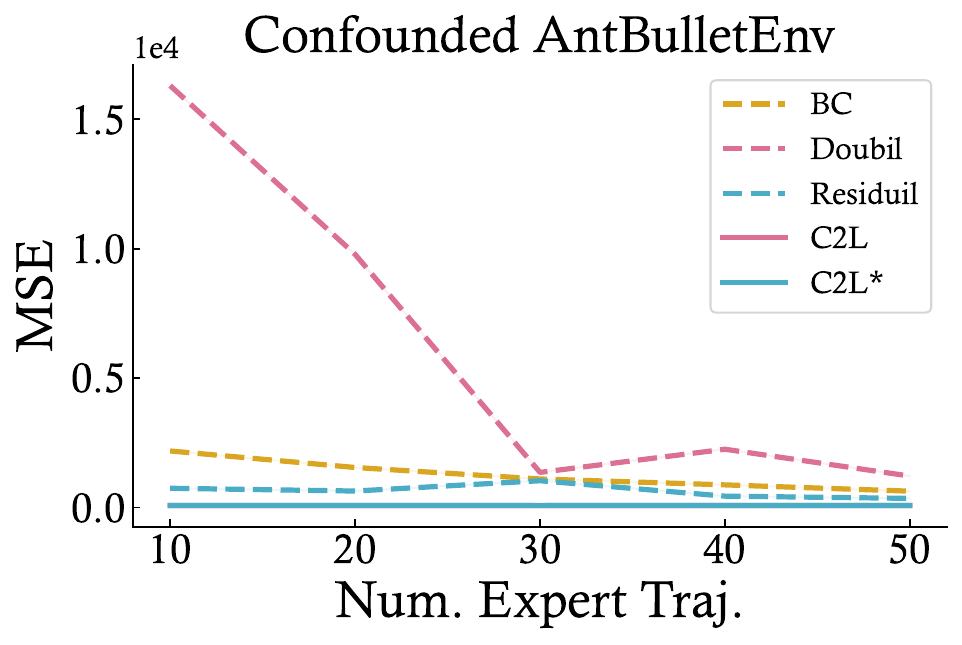}}
    \subfigure{
    \includegraphics[width = 0.32\textwidth]{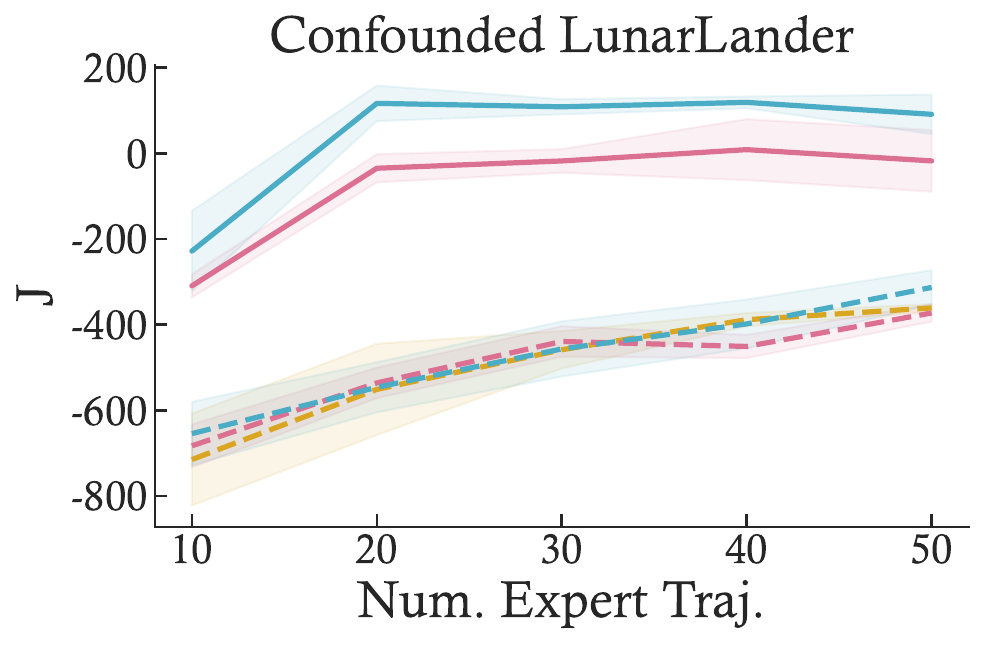}}
    \subfigure{
    \includegraphics[width = 0.32\textwidth]{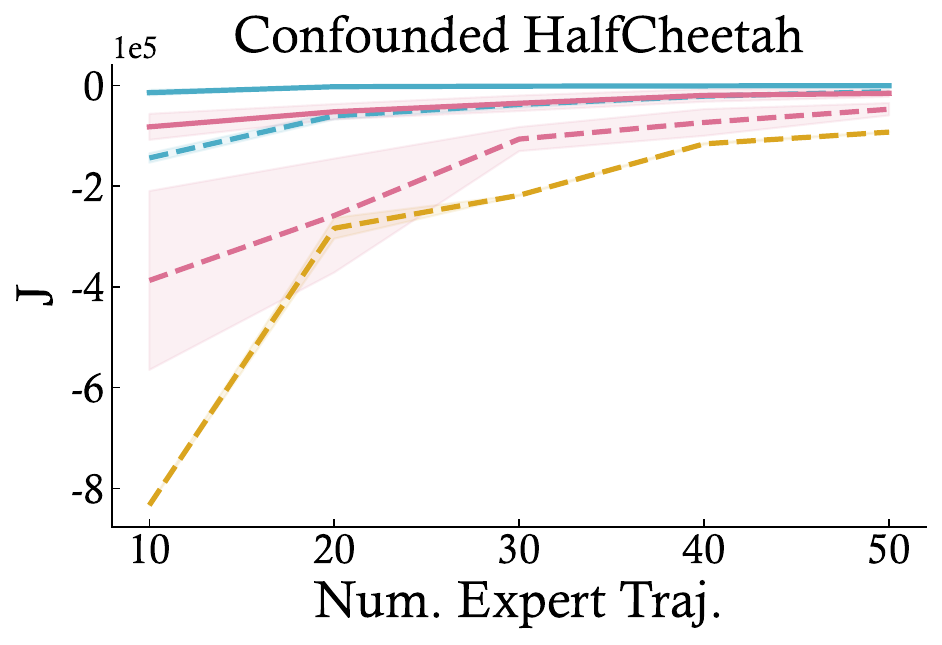}}
    \subfigure{
    \includegraphics[width = 0.32\textwidth]{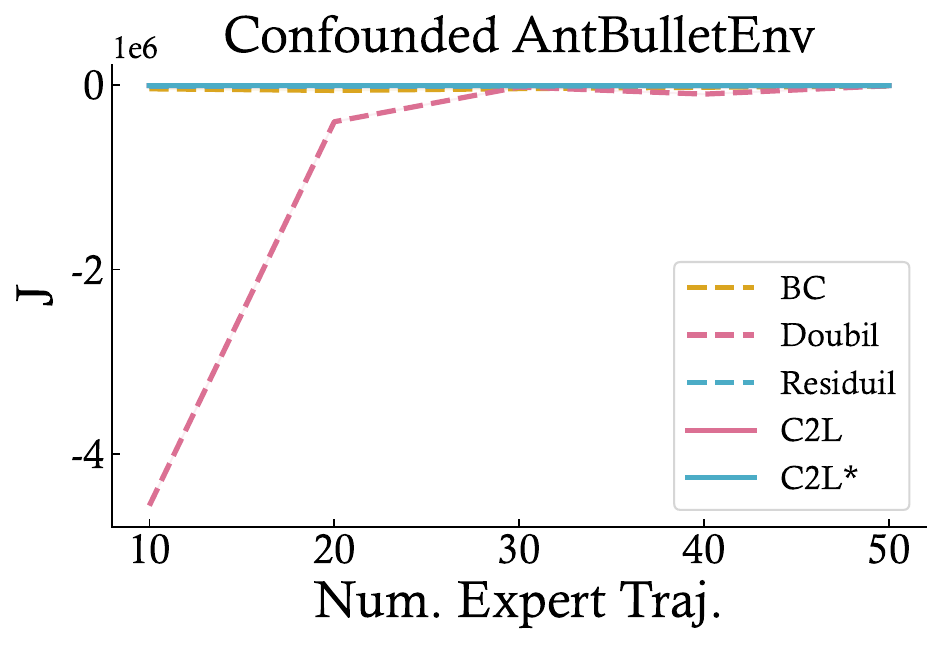}}
    \caption{Experimental results of $\mathrm{MSE}$ and $\mathrm{J}$ for evaluating the learned policy in three confounded environments, LunarLander, HalfCheetah, and AntBulletEnv. 
    The shaded regions represent one standard deviation about the mean across five random seeds.
    We see that both of our approaches, C2L and the C2L*, outperform baselines on all metrics. }
    \vspace{-5mm}
    \label{fig:mse_j in three env}
\end{figure}

We compared our methods, C2L with simulator access and C2L* for offline policy learning against behavioral cloning (BC)~\citep{codevilla2019exploring} and the state-of-the-art confounded imitation learners (ResiduIL and DoubIL)~\citep{swamy2022causal}. We employed two principal metrics: mean squared error ($\mathrm{MSE}$), measuring the L2 distance between predicted and expert actions,
and policy return ($\mathrm{J}$) representing the average episode reward normalized by expert performance. 

\begin{figure}[thb!]
    \centering
    \subfigure{
    \includegraphics[width = 0.32\textwidth]{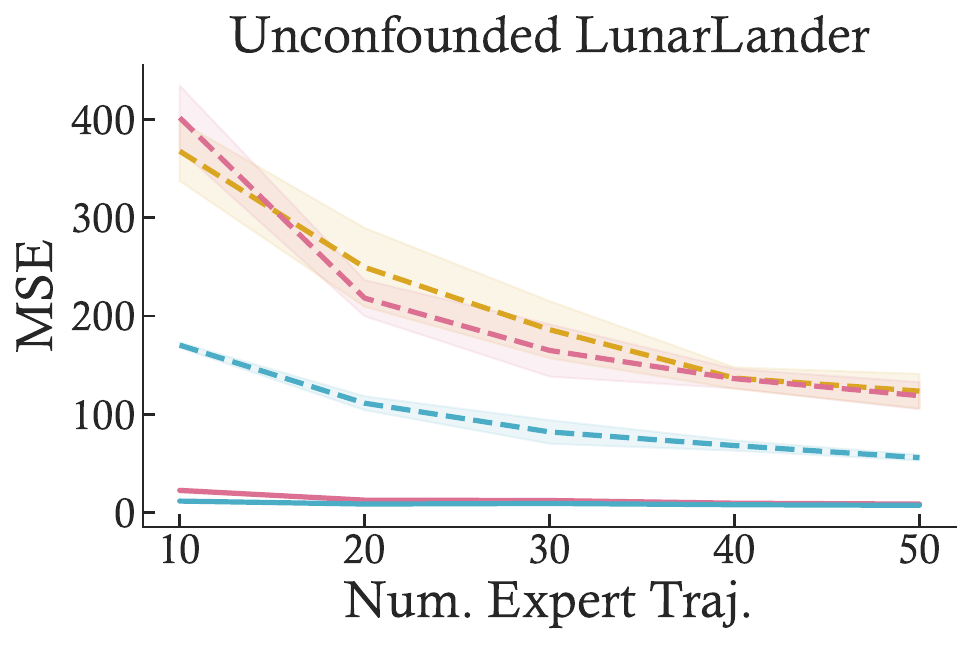}}
    \subfigure{
    \includegraphics[width = 0.32\textwidth]{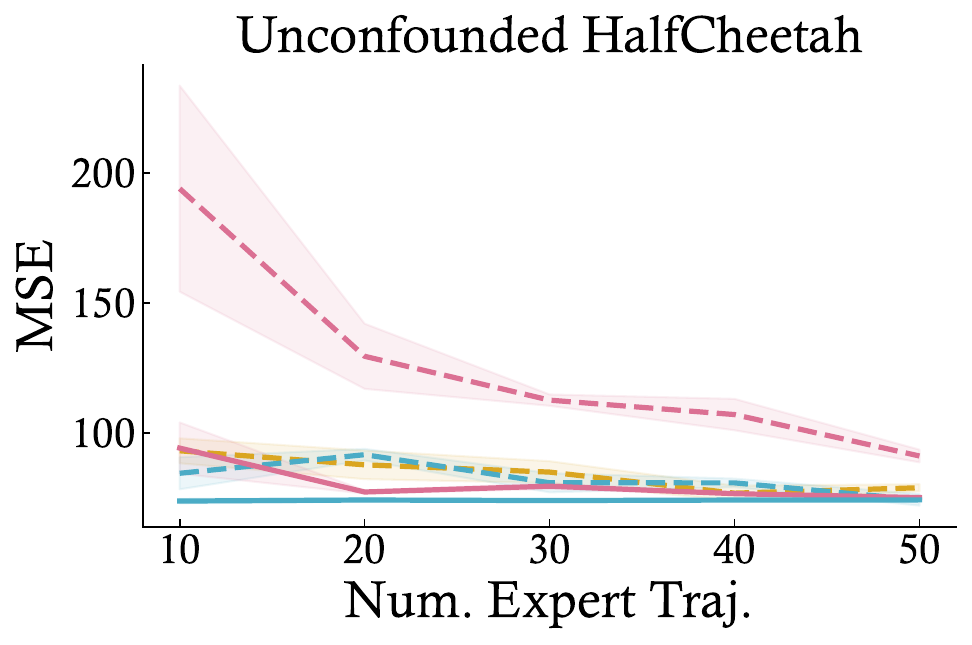}}
    \subfigure{
    \includegraphics[width = 0.32\textwidth]{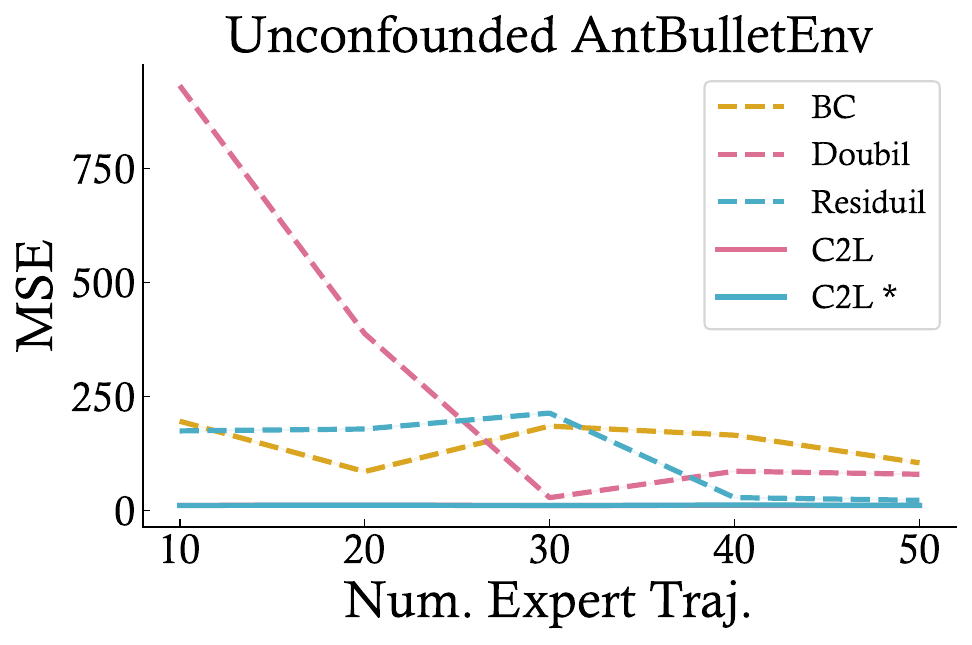}}
    \subfigure{
    \includegraphics[width = 0.32\textwidth]{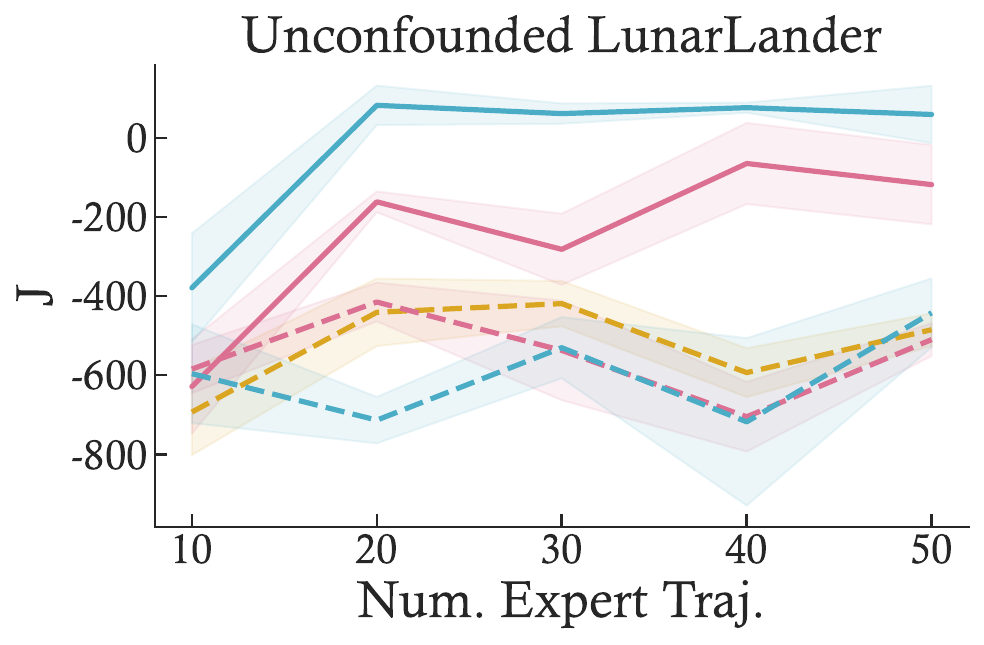}}
    \subfigure{
    \includegraphics[width = 0.32\textwidth]{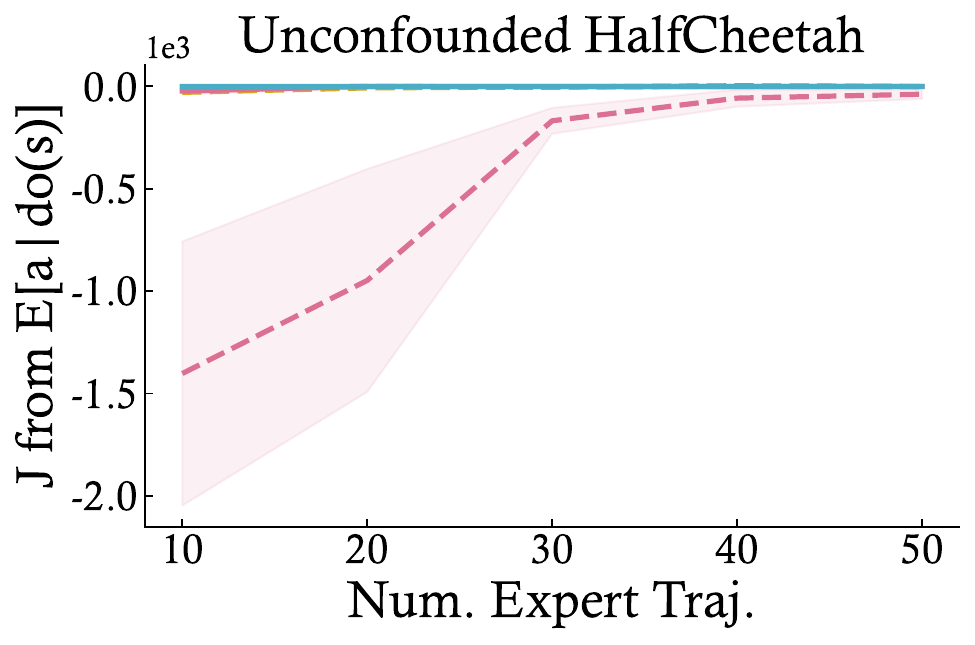}}
    \subfigure{
    \includegraphics[width = 0.32\textwidth]{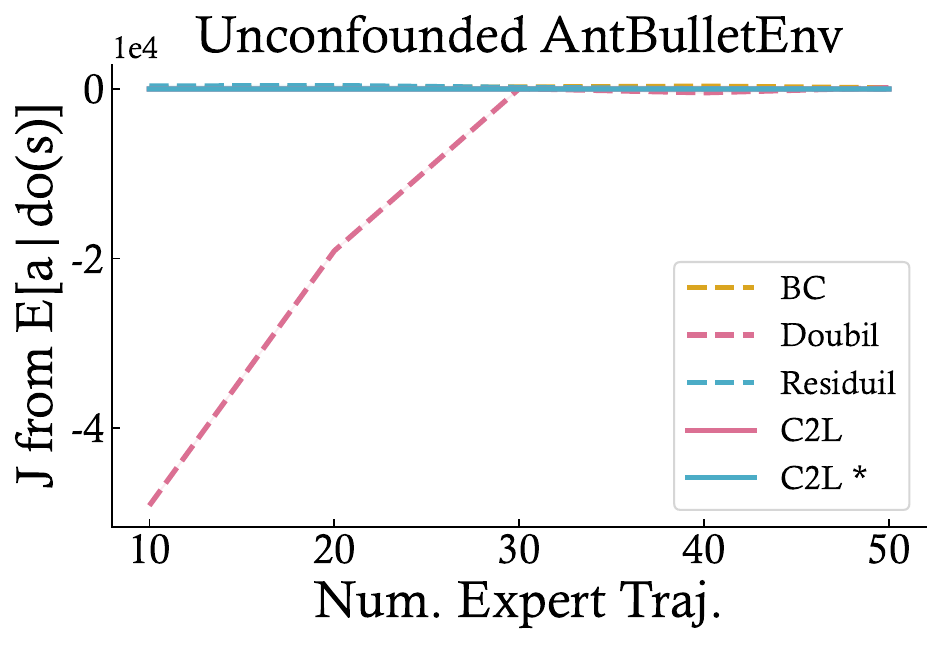}}
    \caption{Experimental results of $\mathrm{MSE}$ and $\mathrm{J}$ for evaluating the learned policy in three unconfounded environments, LunarLander, HalfCheetah, and AntBulletEnv. 
    The shaded regions represent one standard deviation about the mean across five random seeds.
    We see that both of our approaches, C2L and the C2L*, outperform baselines on all metrics. 
    }
    \label{fig:noiseless mse_j in three env}
\end{figure}

As shown in Fig.\ref{fig:mse_j in three env}, we see that while all methods exhibit asymptotic convergence with increasing expert trajectories, our approaches, C2L and C2L*, consistently achieve superior performance over BC, ResiduIL, and DoubIL across all environments and metrics.  Notably, the performance advantage is particularly pronounced in data-scarce regimes, where our methods maintains robust performances even with limited demonstration data.
Further, we conducted experiments in the unconfounded settings. 
%
In Fig.\ref{fig:noiseless mse_j in three env}, we find overall our proposed algorithms, C2L and C2L*, significantly outperform over BC, ResiduIL, and DoubIL, across all environments and metrics. 
Espeically in HalfCheetah and AntBulletEnv, with 50 expert trajectories, most methods could achieve convergence in $\mathrm{MSE}$ and $\mathrm{J}$. Surprisingly, when the number of expert trajectories is small, our methods exhibit a particularly wider performance gap than the baselines.



\section{Conclusion} \label{sec:conclu}
We presented a Confounded Causal Imitation Learning (C2L) model, and derived a two-stage framework to identify the valid IV and learn the policy with no confounding bias. 
We proposed the Auxiliary-Based testing Criterion, namely AB Criterion, with which we showed that the valid IV is identifiable sufficiently and necessarily under appropriate assumptions. 
We also developed two policy optimization approaches, one simulator-based and the other fully offline, that leveraged identified IVs to recover unbiased policies. 
Extensive experiments demonstrated that our algorithms not only accurately select valid instruments across diverse environments and confounding regimes but also outperform existing methods in policy learning. 
While our method effectively handles persistent confounders with time-invariant influence mechanisms, its performance may degrade when facing time-varying confounding patterns. It is an important direction for future research to address non-stationary real-world environments.

\bibliographystyle{plainnat} 
\bibliography{refer}

\end{document}